\documentclass[12pt]{article}

\usepackage{scrextend}
\usepackage{amssymb}  
\usepackage[english]{babel}
\usepackage{graphicx}
\usepackage{epstopdf}
\usepackage{dsfont}
\usepackage{amsmath}
\usepackage[ruled]{algorithm2e}
\usepackage{enumitem}
\usepackage[margin=1in]{geometry}

\begin{document}
	\title{\textbf{Who wins the Miss Contest for Imputation Methods?  Our Vote for Miss BooPF. }}

	\author{Burim Ramosaj\thanks{Ulm University, Institute of Statistics, Ulm, Germany.\newline \text{ \hspace{0.42cm}}  Mail: burim.ramosaj@uni-ulm.de, markus.pauly@uni-ulm.de }  \text{\hspace{0.0001cm}}    and  Markus Pauly\footnotemark[1]}
	
	\maketitle
	
	\begin{abstract}
		Missing data is an expected issue when large amounts of data is collected, and several imputation techniques have been proposed to tackle this problem. Beneath classical approaches such as MICE, the application of Machine Learning techniques is tempting. Here, the recently proposed \texttt{missForest} imputation method has shown high imputation accuracy under the Missing (Completely) at Random scheme with various missing rates. In its core, it is based on a random forest for classification and regression, respectively. In this paper we study whether this approach can even be enhanced by other methods such as the stochastic gradient tree boosting method, the C5.0 algorithm or modified random forest procedures. In particular, other resampling strategies within the random forest protocol are suggested. In an extensive simulation study, we analyze their performances for continuous, categorical as well as mixed-type data. Therein, MissBooPF, a combination of the stochastic gradient tree \textbf{boo}sting method together with the \textbf{p}arametrically bootstrapped random \textbf{f}orest method, appeared to be promising. Finally, an empirical analysis focusing on credit information and Facebook data is conducted.  
	\end{abstract}
	\textbf{Keywords:} Random Forest, Stochastic Gradient Tree Boosting, Resampling, Imputation, MICE
	
	\section{Introduction}
	
	In quantitative science, partially observed data are essential components during the data collection process. Excluding observations with missing values from further analyses could lead to serious information losses. Imputation is one possible solution to tackle this issue by imputing missing values with reasonable estimates and conduct the statistical analyzes as if there has not been any missing values. So far, several imputation methods have been proposed ranging from simple mean imputation to more complex and advanced imputation techniques, see, e.g. \cite{little2014statistical}. In contrast to constant accreditation like mean or median imputation, where dependency structures within the data set are not fully taken into consideration, regression and classification based techniques have prevailed in practice. For example the \textsf{R}-package \texttt{mice} assumes that the data generating process originates from a parametric distribution and imputed values are generated in a cyclic fashion using Gibbs Sampling resulting into chained equations, cf. \cite{van2011multiple} for details. 
	However, extending its usage to more complex data structures, especially when the assumption of a parametric probability distribution is not met, the \texttt{mice} package can lose in imputation accuracy. Recursive partitioning methods, especially regression and classification trees are able to deal with mixed-type data sets while dissociating from the parametric assumption of the data generating process. Advantages such as the capability of dealing with collinearity effects, detecting complex interaction effects, process high dimensional data sets, rare overfitting problems and increased prediction accuracy made them popular in theory and practice.   
	One famous approach of regression and classification trees is the CART algorithm developed in \cite{breiman1984classification}. Several modifications have been proposed such as randomized split selection, bootstrap aggregation (bagging) and boosting (see \cite{hastie2009overview}). In particular, the incorporation of randomness in the CART algorithm has shown favorable prediction accuracy effects, resulting into the random forest algorithm (cf. \cite{breiman2001random}).\\In order to use the benefits of CART-based decision trees, \cite{stekhoven2011missforest,stekhoven2011using} proposed and implemented the \textsf{R} package \texttt{missForest} for predicting missing values
	using the random forest algorithm. As analyzed in \cite{Weljee2013} and \cite{stekhoven2011missforest}, the \texttt{missForest} algorithm has beaten currently available imputation methods as MICE or $k$-nearest neighbor with regard to accuracy. The latter has been measured in terms of normalized root mean squared error and the proportion of false classification. Especially, under the missing completely at random scheme, favorable imputation results could be achieved.  However, a proper analysis of the imputation accuracy under various missing mechanisms was not conducted. Therefore, one part of our simulation study covers the analysis of imputation schemes under the missing completely at random as well as the missing at random mechanism. Furthermore, we were interested in potential modifications towards existing CART-based decision trees such that prediction accuracy can be increased. For this reason, several methods have been taken into consideration, that have not yet been compared to the \texttt{missForest}: C5.0, stochastic gradient tree boosting as well as modifications of the random forest method. The latter benefits by introducing bagging procedures during tree construction, which is limited to simple resampling strategies such as with and without replacement. However, these resampling strategies might be too costly in terms of computational memory and time, when the training data is too large. Therefore,  \cite{Provost1999}  proposed an efficient sampling schedule that does not use  the whole training data. In case of binary response variable, \cite{Weiss2003} analyzed the effect of the class distribution on the tree induction. Since the random forest is constructed with the help of iterative bootstrapping (non-parametrically using simple resampling procedures such as with and without replacement), it is also worth to investigate the effect of other resampling strategies. Here, certain asymptotic models or parametric bootstrap strategies are of specific interest as e.g. proposed by \cite{krishnamoorthy2010,xu2013,Kon:2015} or \cite{smaga2017bootstrap} for continuous MANOVA. Therein, computational time cost and memory can be saved during tree construction, since training data information is compressed into the estimation of bootstrap parameters. Appropriately, we will propose several resampling methods and modify the protocol of the random forest algorithm which may result in increased imputation accuracy.

	\section{Missing Mechanism}
	
	Let $\mathbf{Y}_{i} = (Y_{ij})_{j = 1}^{p} =  (Y_{i1}, ....,Y_{ip})^{\top}$, $i = 1,...,n,$ be independent and identically distributed $p$ - dimensional random vectors with common distribution $\mathbf{F}=(F_j)_{j=1}^p$. The marginal distributions  $F_{j}(t) = \mathbb{P}(Y_{1j} \le t)$, $ j= 1,\dots, p$ are assumed to be either continuous or finitely discrete; corresponding to continuous or categorical outcome variables, respectively. According to this difference, we divide $\{1,...,p \} = \mathcal{C}_{1} + \mathcal{C}_{2}$ into continuous $(\mathcal{C}_{1})$ and categorical $(\mathcal{C}_{2})$ components, respectively. Let $\mathbf{Y} = (Y_{ij})_{i,j} =  (\mathbf{Y}_{1}...\mathbf{Y}_{n})^\top \equiv (\mathbf{X}_{1}...\mathbf{X}_{p}) \in \mathbb{R}^{n \times p}$ denote the corresponding data matrix and $\mathbf{R} = (R_{ij})_{i,j}$ 
	$\in \{0,1\}^{n \times p}$ the matrix indicating whether $Y_{ij}$, $i = 1,...,n$, $ j = 1,...,p$ is observed $(R_{ij} = 1)$ or not $(R_{ij} = 0)$. Further, let $\mathbf{Y}_{obs}$ and $\mathbf{Y}_{mis}$ be the observed and missing parts of $\mathbf{Y}$. In \cite{rubin1976inference} the missing mechanism is defined through a probabilistic model, where $\mathbf{R}$ depends on some unknown parameter $\xi$ and the random matrix $\mathbf{Y}$. It is said to be 
	\begin{itemize}
		\item \textit{Missing completely at random} (MCAR), if\\ $\mathbb{P}( \mathbf{R} | \mathbf{Y}, \xi ) = \mathbb{P}(\mathbf{R}| \xi). $
		\item  \textit{Missing at random} (MAR), if \\
		$ \mathbb{P}( \mathbf{R} | \mathbf{Y}, \xi ) = \mathbb{P}(\mathbf{R}|\mathbf{Y}_{obs}, \xi). $   
		\item \textit{Missing not at random} (MNAR), if \\
		$ \mathbb{P}( \mathbf{R} | \mathbf{Y}, \xi ) = \mathbb{P}( \mathbf{R} | \mathbf{Y}_{obs}, \mathbf{Y}_{mis}, \xi ) \neq \mathbb{P}(\mathbf{R}|\mathbf{Y}_{obs}, \xi). $       
	\end{itemize}
	
	In quantitative science, mainly two approaches have been established to handle missing data. Imputation and data adjusted methods. For the latter the missing mechanism is incorporated in the likelihood of the data analysis model resulting into parametric or non-parametric MLE-type methods, see e.g. \cite{vach1994missing,schafer1997analysis,xu2012accurate,konietschke2012ranking,amro2017permuting}. In contrast, imputation resp. multiple imputation assigns each missing value a reasonable estimate such that partially observed data is completed and the analysis can be conducted as if there has not been any missing values. Latter incorporates the uncertainty of the imputation procedure into sample estimates by predicting several values. \cite{little2014statistical} proved that sample estimates remain unaffected in case of an MCAR mechanism. However, for regression and classification trees, MCAR can result into biased estimates, as \cite{strobl2007unbiased} have counteracted. Therefore, increasing imputation accuracy by imputing missing values close to the real data under MCAR or MAR mechanisms is an essential gain in later statistical inference. In a comparative simulation study, we want to illuminate the predictive accuracy of various imputation techniques under these missing mechanisms, by artificially inserting missing values in synthetic and empirical data sets. 
	
	\nocite{hapfelmeier2014new}
	
	\section{Methodology}\label{sec:Meth}
	
	In this section, we give a brief introduction of CART-based methods by enlightening the random forest, the C5.0 as well as the stochastic gradient tree boosting method. For the MICE method, we refer to \cite{van2011multiple}. 
	
	Tying to the previous notation, we denote with 
	$\mathbf{Y}_{\pi} = (Y_{i\pi(j)})_{i,j} = (\mathbf{X}_{\pi(1)}...\mathbf{X}_{\pi(p)}) \in \mathbb{R}^{n \times p}$
	the column-per- muted data matrix for a permutation $\pi: \{1,...,p\} \rightarrow \{ 1,...,p\}$. 
	Furthermore, let  $\mathbf{i}_{\pi(j)}^{obs} = \{ i \in \{1,...,n\} | R_{i\pi(j)} = 1 \}$ be the index set of observed components of its $j$-th column $\mathbf{X}_{\pi(j)}$ and denote the corresponding $|\mathbf{i}_{\pi(j)}^{obs}|$-dimensional sub-vector of observed components as $\mathbf{X}_{\pi(j)}^{obs}$.
	Accordingly, we define $\mathbf{i}_{\pi(j)}^{mis} = \{ i \in \{1,...,n\} | R_{i \pi(j)} = 0  \}$ as the set of indices, where the components of $\mathbf{X}_{\pi(j)}$ are missing and $\mathbf{X}_{\pi(j)}^{mis}$ the sub-vector with missing components 
	in $\mathbf{X}_{\pi(j)}$. In addition, let $
	\mathbf{Y}_{- \pi(j)}^{obs} = (Y_{i\pi(1)},\dots, $ $Y_{i\pi(j-1)}, Y_{i\pi(j+1)},\dots,Y_{i\pi(p)} )_{i \in \mathbf{i}_{\pi(j)}^{obs} }
	$ be the permuted data matrix without the random vector $\mathbf{X}_{\pi(j)}$ and all entries, where $\mathbf{X}_{\pi(j)}$ has observed components. Hence, we set $\mathbf{Y}_{- \pi(j)}^{mis} = (Y_{i\pi(1)},\dots,Y_{i\pi(j-1)},$ 
	$Y_{i\pi(j+1)},$ $\dots,Y_{i\pi(p)} )_{i \in \mathbf{i}_{\pi(j)}^{mis} }$, $j = 1,...,p$ as the permuted data matrix without $\mathbf{X}_{\pi(j)}$ but with all other entries where $\mathbf{X}_{\pi(j)}$ has missing components. Then, the general  algorithm of \cite{stekhoven2011missforest} for imputing missing values within the class of decision-tree based regression and classification algorithms can be summarized as
	\begin{algorithm}
		\SetKwInOut{Input}{Input}
		\SetKwInOut{Output}{Output}
		\SetKw{Steps}{Steps}
		\Input{Data matrix $\mathbf{Y}$ with missing values.}
		\Output{Imputed data matrix $\mathbf{Y}^{imp}$.}
		\vspace{0.25cm}
		\Steps: \\
		\begin{enumerate}[label=\textbf{\arabic*}:]
			\item Sort the $p$ columns of the data matrix $\mathbf{Y}$ based on the missing rate in ascending order resulting into a permutation $\pi(1),...,\pi(p)$ of the $p$ outcome variables. 
			\item Initially impute mean resp. mode values for missing continuous resp. categorical realizations of random variables. 
			\item Starting with the first permuted column $X_{\pi(1)}$, separate the permuted data matrix  $\mathbf{Y}_{\pi}$ into four parts using the previous notation:
			$$  
			\left[ \begin{array}{c|c}
			\mathbf{X}_{\pi(1)}^{obs} & \mathbf{Y}_{-\pi(1)}^{obs} \\
			\hline
			\mathbf{X}_{\pi(1)}^{mis} & \mathbf{Y}_{-\pi(1)}^{mis} 
			\end{array} \right] \in \mathbb{R}^{ (| \mathbf{i}_{\pi(1)}^{obs} | + | \mathbf{i}_{\pi(1)}^{mis} |) \times (1 +  (p-1)) }. $$
			\item Train the chosen regression resp. classification tree  method using the sub-matrix $\mathbf{Y}_{-\pi(j)}^{obs}$  as covariates and $\mathbf{X}_{\pi(j)}^{obs}$ as response variable. 
			\item Impute the missing values of $\mathbf{X}_{\pi(j)}$ using the trained regression resp. classification model with $\mathbf{Y}_{- \pi(1)}^{mis}$ as covariate values. Then move to the next variable and repeat steps $\mathbf{3}$ and $\mathbf{4}$ until all variables have been treated.
			\item As long as the $L_2$-error of the newly and previously imputed data set has not increased for the first time, return to step $3$. 
		\end{enumerate}
		\caption{Missing Value Imputation.}
	\end{algorithm}
	
	Thus, the algorithm turns the missing value problem into a regression or classification problem, depending on the scale of the response variable. In this way, the CART-based procedure tries to overcome the initial issue of an unsupervised learning problem, by training the algorithm on observed values and predicting missing values afterwards. Recall that this does not necessarily lead to an MAR-respecting mechanism, since being observed is defined through $\mathbf{i}_{\pi(j)}^{obs}$, i.e. $\mathbf{Y}_{-\pi(j)}^{obs}$ may have a missing value in a variable other than $\pi(j)$. CART-based  ensemble methods such as the random forest or the stochastic gradient tree boosting algorithm, however, benefit from the construction of several base learners and their later combination by majority vote or averaging. Their combination and construction of trees nevertheless differ as described below: 
	\begin{enumerate}
		\item \textbf{Random Forest:}\\ For each decision tree, non-parametric bootstrap sampling is performed selecting $N \le n$ observations either with or without replacement from the training set. In case of selecting without replacement, $63.2 \%$ of the observations are selected for training. Furthermore, feature subspacing at each terminal node is conducted shrinking the potential number of split variables to $m \in \{1,...,p\}$. For classification, $m = \sqrt{p}$ is chosen, whereas for regression, $m = p/3$ is selected. The choices are motivated by a simulation study conducted in \cite{breiman2001random} and the trade-off between computational time and prediction accuracy (cf. \cite{stekhoven2011using}). 
		\item \textbf{Stochastic Gradient Tree Boosting:}\\ The trees are constructed in sequential order by minimizing prediction error with respect to a loss function, say $\psi$. It is chosen depending on the nature of the variable: For continuous outcomes, the squared error loss is used while for categorical outcomes, either the Bernoulli loss or the multinomial loss is used subject to level size. Minimization is conducted using the gradient descent method. The subsequent tree is fitted to a bootstrap sample (without replacement) of $\mathbf{Y}_{-\pi(j)}^{obs}$ of size $N \le n$ and its corresponding pseudo-residuals of the previous (additive) tree as training signal. The final decision tree is then constructed by taking the weighted sum of all trees.   
		\item \textbf{C5.0:}\\
		The predecessor of the classification algorithm C5.0 has been considered as the best among $10$ data mining algorithms by the IEEE
		International Conference on Data Mining (ICDM) in December $2006$. Since the implemented version of the algorithm in \textsf{R} is only capable of solving classification tasks, in our scheme of work, C5.0 is used for imputing categorical outcomes only. The construction of trees is similar to the general CART method. In contrast to the random forest method, C5.0 does not conduct feature subspacing and it is able to construct multiway split trees while using the Shannon entropy as impurity measure. Motivated by \cite{Friedman2002}, the algorithm is able to conduct boosting similar to the stochastic gradient tree boosting. In addition, pessimistic pruning is executed and decision trees can be transformed into rulesets (cf. \cite{C5.0Buljow}). 
	\end{enumerate} 
	
	Stochastic gradient tree boosting and C5.0, however, have not yet been analyzed within the current missing framework and their imputation accuracy using the scheme of algorithm $1$. Although the stochastic gradient tree boosting method and the C5.0 differ technically, their approach is based on the same boosting principle. Since its technical details are rather undocumented, missing values using the C5.0 algorithm are imputed similarly to the stochastic gradient tree boosting for categorical response variables.
	
	To explicitly explain the differences between the approaches denote with $ \{\Gamma(\mathbf{x}; \Theta_{b}) \}_{b=1}^{B}$ a collection of $B$ decision trees with random seed $\Theta_{b}$, specifying the nature of each constructed tree and input vector $\mathbf{x} \in \mathbb{R}^{p-1}$. Then for $i \in \mathbf{i}_{\pi(j)}^{mis}, j = 1,\dots,p,$ the random forest algorithm imputes via
	
	$$ \hat{Y}_{i \pi(j)}^{RF} = \begin{cases}
	\frac{1}{B} \sum\limits_{b = 1}^{B} \hat{\Gamma}(\mathbf{Y}_{i,-\pi(j)}^{mis}; \Theta_{b}), & \mbox{$\pi(j) \in \mathcal{C}_{1} $ ,} \\
	\text{Mo}(\hat{\Gamma}(\mathbf{Y}_{i,-\pi(j)}^{mis}); \Theta_{b})), & \mbox{$\pi(j) \in \mathcal{C}_{2} $ ,}
	\end{cases} $$
	where $\hat{\Gamma}$ is the trained ensemble of decision trees on $(\mathbf{Y}_{-\pi(j)}^{obs},\mathbf{X}_{\pi(j)}^{obs})$ and $\text{Mo}(\mathbf{b})$ denotes the modus of a vector $\mathbf{b}$. For the stochastic gradient tree boosting, imputation is conducted as follows: 
	
	$$ \hat{Y}_{i \pi(j)}^{GB} = \begin{cases}
	\sum\limits_{b = 1}^{B} \beta_{b} \hat{\Gamma}( \mathbf{Y}_{i, -\pi(j)}^{mis} ; \Theta_{b}), &\text{} \pi(j) \in \mathcal{C}_{1}, \\
	\arg\max\limits_{k} \hat{\Gamma}_{k}(\mathbf{Y}_{i, -\pi(j)}^{mis};\Theta_{b} ), &\text{} \pi(j) \in \mathcal{C}_{2},
	\end{cases}
	$$
	
	where $\hat{\Gamma}_{k}(\mathbf{Y}_{i, -\pi(j)}^{mis},\Theta ) = \sum\limits_{b = 1}^{B} \beta_{b}^{(k)} \hat{\Gamma}( \mathbf{Y}_{i, -\pi(j)}^{mis} ; \Theta_{b}^{(k)}) $ estimates the class probability $\mathbb{P}(Y_{i\pi(j)} = k | \mathbf{Y}_{- \pi(j)} )$ in case of a categorical variable with $k \in dom( \mathbf{X}_{\pi(j)})$. The weights $\{ \beta_{b}\}_{b = 1}^{B}$ resp. $\{ \beta_{b}^{(k)}\}_{b = 1}^{B}$, with $k \in dom( \mathbf{X}_{\pi(j)})$ are chosen by iteratively minimizing $$\sum\limits_{i = 1}^{N} \psi[Y_{i\pi(j)}, \sum\limits_{l = 1}^{b-1} \beta_{l}\hat{\Gamma}(\mathbf{Y}_{i, -\pi(j)}^{obs} ; \Theta_{l})+ \beta_{b^*}\hat{\Gamma}( \mathbf{Y}_{i, -\pi(j)}^{obs} ; \Theta_{b})  ]$$  with respect to $\beta_{b^*}$, $b^* = 1,...,B$ (cf. \cite{Friedman2002}).  The prediction of missing values using the C5.0 algorithm is conducted in a similar fashion to the stochastic gradient tree boosting, where the difference lies in the construction of each single tree by using the Shannon entropy and pessimistic pruning. Although the stochastic gradient tree boosting and the random forest algorithm both make use of regression and classification trees, they mainly differ in the way the trees are constructed. Since stochastic gradient tree boosting fits subsequent trees on pseudo-residuals, in principle, it is not possible to implement this algorithm in a parallel fashion. Furthermore, their focus in achieving higher prediction accuracy is different. Recalling the decomposition of the mean squared error loss function evaluated in $\mathbf{x} \in \mathbb{R}^{p-1}$ into 
	$$Var_{\Theta}(\hat{\Gamma}(\mathbf{x};\Theta_{b})) + Bias_{\Theta}(\hat{\Gamma}(\mathbf{x};\Theta_{b}))^{2}, $$ 
	the random forest aims to achieve its decrease in mean squared error by the construction of (de-correlated) trees resulting into the decrease through variance reduction:  
	
	\begin{align*}
	Var_{\Theta}(\frac{1}{B}\sum_{b = 1}^{B}\hat{\Gamma}(\mathbf{x};\Theta_{b})) &= \frac{1 - \rho(\mathbf{x})}{B}Var_{\Theta}(\hat{\Gamma}(\mathbf{x};\Theta)) \\ &+ 
	\rho(\mathbf{x}) Var_{\Theta}(\hat{\Gamma}(\mathbf{x};\Theta)).
	\end{align*}
	Here, $\rho(\mathbf{x}) \in [-1,1]$ is the pairwise tree correlation at $\mathbf{x} \in \mathbb{R}^{p-1}$. On the other hand, stochastic gradient tree boosting aims to minimize mean squared error by decreasing potential bias. This is realized through fitting subsequent trees not to the training signal $\mathbf{X}_{\pi(j)}^{obs}$, $j = 1,...,p$,  but to the pseudo-residuals obtained from the previous (additive) tree model. However, both methods require to find an optimal value of the tree size $B$ within the missing scheme. In addition, the stochastic gradient tree boosting minimizes a loss function $\psi$, which in practice is conducted using gradient descent method. Hence, the choice of the step-size in the direction of the steepest descent clearly affects prediction accuracy. 
	
	\section{New Resampling Proposal}\label{sec:Res}
	In this section, we propose modifications of the resampling technique within the random forest method. We extend the possibility of constructing CART-based ensemble trees from simple random sampling to parametric sampling. Since we request to treat both, continuous and categorical variables, several sampling strategies are considered: 
	
	\begin{enumerate}
		\setlength{\itemsep}{3pt}
		\item \textbf{Stratification:} Assuming that the data set consists of ordinal or nominal variables, a stratified sampling procedure with replacement is conducted. This way, we aim to represent low frequent levels of the response variable into the training phase of the learning algorithm accordingly. The latter would, e.g. not be the case when the resample is generated from a multinomial distribution (results not shown).
		\item \textbf{Multivariate Normal Distribution:} If the data matrix $\mathbf{Y}$ consists of only continuous realizations, then the bootstrap sample in the random forest procedure is substituted by a so-called asymptotic-model based (or parametric) bootstrap sample generated from $N_p(\widehat{\boldsymbol{\mu}}, \widehat{\mathbf{\Sigma}})$. Here, the mean vector $\boldsymbol{\mu}$ and the covariance matrix $\mathbf{\Sigma}$ are estimated by their empirical counterparts $\widehat{\boldsymbol{\mu}}$ and $\widehat{\mathbf{\Sigma}}$ calculated from $(\mathbf{Y}_{- \pi(j)}^{obs}, \mathbf{X}_{\pi(j)}^{obs})$, $\pi(j) \in \{1,...,p\}$.  
		\item \textbf{Multivariate Kernel:} Similar to the latter case we additionally propose to draw the resample from a kernel estimator $\hat{f}$ for multivariate data. In its general form it is given by
		$$ \hat{f}_{\mathbf{Y}}(\mathbf{y}; \mathbf{H}) = \frac{1}{n} \sum\limits_{i = 1}^{n} K_{\mathbf{H}}(\mathbf{y} - \mathbf{Y}_{i} ), \quad \mathbf{y} \in \mathbb{R}^{p}, $$
		where $\mathbf{H}$ is the bandwidth matrix and $K_{\mathbf{H}}(\mathbf{y}) = |\mathbf{H}|^{-1/2} K(\mathbf{H}^{-1/2}\mathbf{y})$. The fitting quality towards the sample $\mathbf{Y}$ is more an issue of the choice of the bandwidth matrix $\mathbf{H}$ rather than the choice of the kernel function (cf. \cite{Mueller2016}). Hence, we choose the Gaussian kernel $K_{\mathbf{H}}(\mathbf{y}) = (2 \pi)^{-p/2} |\mathbf{H}|^{-1/2}\exp\{$ $ -\frac{1}{2} \mathbf{y}^{T} \mathbf{H}^{-1} \mathbf{y} \}$. The selection of $\mathbf{H}$ is usually conducted by minimizing the mean integrated squared error. Since this minimization problem is generally intractable, several classes of bandwidth matrices have been proposed, see e.g. \cite{Wand1994}. We simply follow the normal scale rule given by 
		$$ \hat{H}_{NS} = \left[\frac{4}{n(d+2)}\right]^{2/(d+4)} \hat{\mathbf{\Sigma}}.$$ 
		As proven in \cite{Chacon2011}, this is the plug-in estimate in case of a normal density $f$ under regularity conditions. 
	\end{enumerate}
	
	\section{Simulation}
	
	In this section we investigate the performance of the algorithms described in Sections~\ref{sec:Meth} and \ref{sec:Res} using synthetic as well as empirical data. Performance is measured by means of the normalized root mean squared error (NRMSE) for continuous variables and proportion of false classification (PFC) for categorical variables, i.e.
	
	$$ NRMSE = \sqrt{\frac{\sum\limits_{j \in \mathcal{C}_1 }  \sum\limits_{i \in \mathbf{i}_{\pi(j)}^{mis}} (Y_{i \pi(j)}^{true} - Y_{i \pi(j)}^{imp} )^{2} }{ \sum\limits_{j \in \mathcal{C}_1} \sum\limits_{i \in \mathbf{i}_{\pi(j)}^{mis}}(Y_{i \pi(j)}^{true} - \bar{Y}_{\pi(j)}^{true})^{2} }}, $$

	$$ PFC =  \frac{1}{\sum\limits_{j \in \mathcal{C}_2}|\mathbf{i}_{\pi(j)}^{mis}|}  \sum\limits_{j \in \mathcal{C}_2} \sum\limits_{i \in \mathbf{i}_{mis}^{\pi(j)}}\mathds{1}\{ Y_{i \pi(j)}^{true}  \neq  Y_{i \pi(j)}^{imp} \}.$$
	
	Here, $Y_{i \pi(j)}^{true}$ with $i \in \mathbf{i}_{\pi(j)}^{mis}$ and $\pi(j) \in \{1,...,p\}$ denotes the $\pi(j)$-th component of the row vector $\mathbf{Y}_{i}$ before missing values have been inserted artificially. Correspondingly, $Y_{i \pi(j)}^{imp}$ is its imputed value. Additionally, we identify with $\bar{Y}_{\pi(j)}^{true}$ the mean value of the sequence $\{ Y_{i \pi(j)}^{true} \}_{i \in \mathbf{i}_{\pi(j)}^{mis} }$, $\pi(j) \in \{1,\dots,p\}$.  
	
	First, PFC resp. NRMSE is measured for synthetic data consisting of only categorical resp. continuous variables. Thereafter, mixed-type data from empirical studies are analyzed. For each of the first two settings a total number of $n = 250$ observations were generated with $p = 15$ variables. For the case of $\mathbf{Y}$ consisting of categorical realizations of random vectors only, the variables where split into $p_{n} = 7$ nominal variables with levels $l_{j} \in\{ A,B,C,D \}, \quad j = 1,\dots,p_{n} $ and $p_{o} = 8$ ordinal variables with $l_{k} \in \{1,2,3,4\}$, $k = 1,\dots,p_{o}$. Missing values are artificially generated under the MCAR and MAR mechanism using a missing rate of $r = 0.1, 0.2$ and $0.3$ with $100$ Monte Carlo runs. Hence, the absolute number of missing observations is not defined through $n$, but through $n \cdot p$, i.e. $\lceil r \cdot n \cdot p \rceil$ is the total number of missing observations. 
	
	For every simulation set-up, the one-sided Brunner-Munzel test \cite{BrunnerMunzel} is conducted in favor of the resulting 'best' procedure, due to its robustness towards heterogeneous settings. Since the random forest as well as stochastic gradient tree boosting are ensemble methods with tree structures as base learners, the choices of hyper-parameters such as tree size and step-size are highly influential on the final prediction accuracy. Due to an additional discretization process for continuous outcomes using the CART algorithm, an increased number of base learners will remarkably increase computational time cost. Hence, following the empirical analysis of \cite{hastie2009overview} for the stochstic gradient tree boosting method and \cite{stekhoven2011missforest} for the random forest method, we set the number of base learners for the random forest to $100$ and for the stochastic gradient tree boosting to $2,000$. Furthermore, the step-size of the gradient descent procedure within  the stochastic gradient tree boosting is set to $0.001$. In contrast to the findings of \cite{breiman2001random} and \cite{stekhoven2011using}, we set the number of split variables to $\sqrt{p}$ for both, regression and classification tasks. A similar approach was also followed by \cite{stekhoven2011missforest}. Finally, note that the p-values of each Brunner-Munzel test are encoded as ${}^* < 0.1$, ${}^{**} < 0.05$ and ${}^{***} <0.01$ over each boxplot.

	\subsection{Categorical variables only}
	We create $p = 15$ correlated multivariate (ordinal) variables with the help of the \textsf{R} function \texttt{rmvord} (cf. \cite{Kaiser2011generating}). The generation is based on thresholding a multivariate normal distribution. However, $p_{n} = 7$ variables are treated as nominal outcomes and their feature presentation is transformed into elements of $\{A, B, C, D\}$. Setting the pairwise correlation $\rho_{kl} = Cor(X_{k}, X_{l})$ to $0.4$, with $k,l \in \{1,\dots,15\}$, we create $15$ ordinal variables and treat the first seven variables in the data matrix $\mathbf{Y}$ as nominal variables. Each of them have four levels: either $A,B,C,D$ or $1,2,3,4$ for nominal, resp. ordinal variables. In order to measure the effect of stratified sampling, two synthetic data sets have been generated: 
	\begin{itemize}
		\item \textbf{(D1)} In the first setting, the relative frequency is simulated using a Dirichlet prior of order $4$ with equal concentration parameters $\alpha_{1}=\dots=\alpha_{4}=100$.  
		\item \textbf{(D2)} For the second data set, a Dirichlet prior of order $4$ with unequal concentration parameters given by  $\boldsymbol{\alpha} = (100,200,500,500)^{T}$ is used for generating relative frequencies.   
	\end{itemize} 
	\begin{figure}[h]
		\centering
		\includegraphics[scale = 0.7]{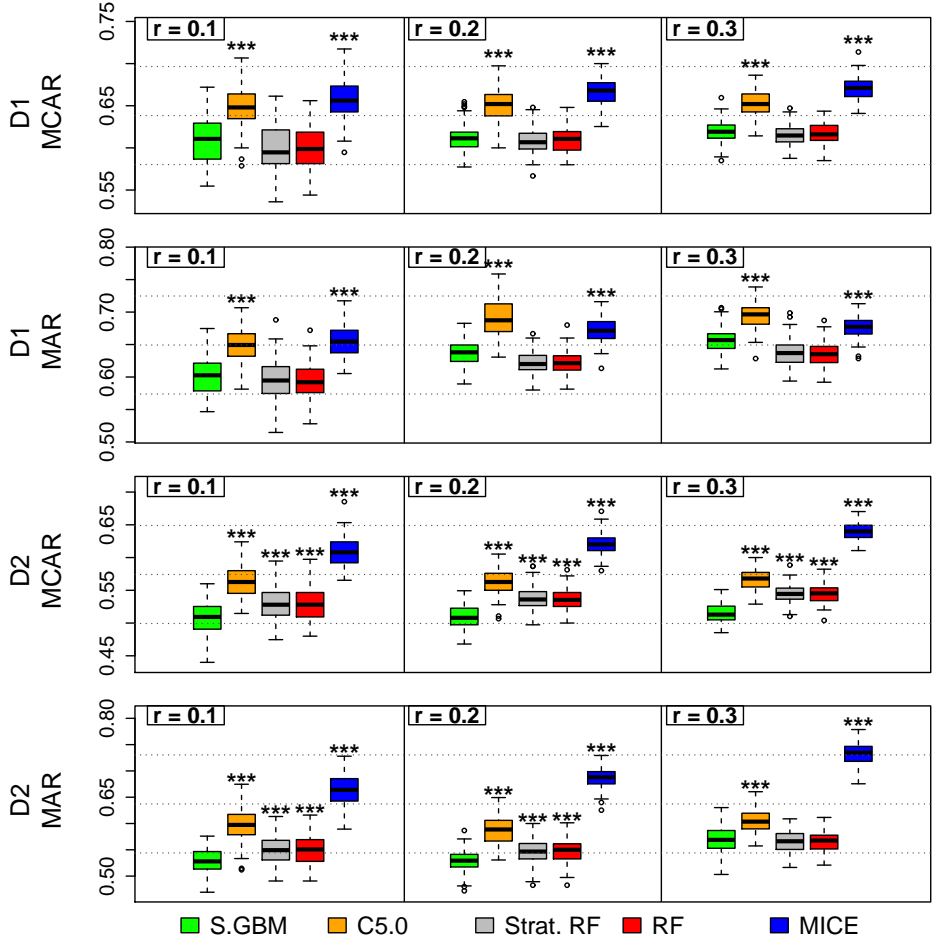}
		\caption{ PFC for data sets \textbf{(D1)} and \textbf{(D2)} under both missing mechanims with various missing rates 
			for the stochastic gradient tree boosting (S.GBM), the C5.0, the random forest with stratified sampling (Strat. RF), the (classical) random forest (RF) and the \texttt{mice} method (MICE).}
		\label{fig1}
	\end{figure}
	
	The simulation results for MCAR and MAR are displayed in Figure \ref{fig1}. They reveal that the novel random forest method with stratified sampling achieved comparable results to the usual random forest method with simple resampling under all twelve simulation settings. Hence, no additional gain in imputation accuracy could be achieved by changing the resampling method to stratification. Although the stochastic gradient tree boosting (S.GBM) method does not distinguish itself compared to random forest (RF) under data set (D1), imputation performance increased if level frequencies are not equally present. Contrary, S.GBM resulted in slightly higher impuation accuracy results compared to the random forest in data set (D2) under the MCAR and MAR scheme. The results were significant at the $1 \%$ level using the one-sided Brunner-Munzel test in favor of the S.GBM.  For example, under the MAR sheme with $20 \%$ missing rate in data set (D2), the mean PFC could be reduced by around $3.7\%$ using the S.GBM. The effect in PFC decrease in the same data set was larger when the missing mechanism was changed to an MAR-respecting procedure leading to a decrease of around $5\%$. The C5.0 method and MICE performed worse for both data sets and missing mechanisms and are clearly dominated by the other three approaches.
	\newpage
	
	\subsection{Continuous variables only}
	
	We generate synthetic data containing only continuous realizations of random variables with $n = 250$ observations originating from the following $p=15$-variate distributions: 
	
	\begin{itemize}
		\item \textbf{(D3)} A multivariate normal $N_{15}(\boldsymbol{\mu}, \boldsymbol{\Sigma})$-distribution with parameter $\boldsymbol{\mu} =   (2,3,...,16)^{\top}$ and covariance matrix $\boldsymbol{\Sigma} = 9\boldsymbol{I}_{15} + 6.3 \boldsymbol{J}_{15}$. Here, $\boldsymbol{J}_{15}$ denotes the $(15\times 15)$ matrix of ones and $\boldsymbol{I}_{15}$ the identity. Note, that this implies a constant correlation coefficient of $\rho_{kl} = 0.7$ for two different variables $X_{k}, X_{l}$, $k,l \in \{1,...,15\}$.
		\item Each observation in the data set is generated from $\boldsymbol{Y}_{i}^{\top} = \boldsymbol{\mu} + \boldsymbol{\Sigma}^{1/2} \boldsymbol{\epsilon}_{i}$ for $i = 1,...,250$, where $\boldsymbol{\mu} = (2,3,...,16)^{\top}$ and 
		$\mathbf{\Sigma} = (\sigma_{kl})_{k,l}$ is chosen such that $\sigma_{kk}= k$ and $\sigma_{kl} / (kl)^{1/2} = 0.7 $ for $k\neq l$, $k, l \in \{1,\dots, 15\}$. We assume further that $\epsilon_{ki}$ are either $iid$ $\chi_{df}^{2}$ distributed with $df = 3$ \textbf{(D4)} resp. $df = 30$ \textbf{(D5)} degrees of freedom or $iid$ log-normally distributed with location parameter $\mu = 0$ and scale parameter $\sigma = 1$ \textbf{(D6)} resp. $\sigma = 2$ \textbf{(D7)}. 
	\end{itemize}
	
	\begin{figure}[h]
		\centering
		\includegraphics[scale = 0.7]{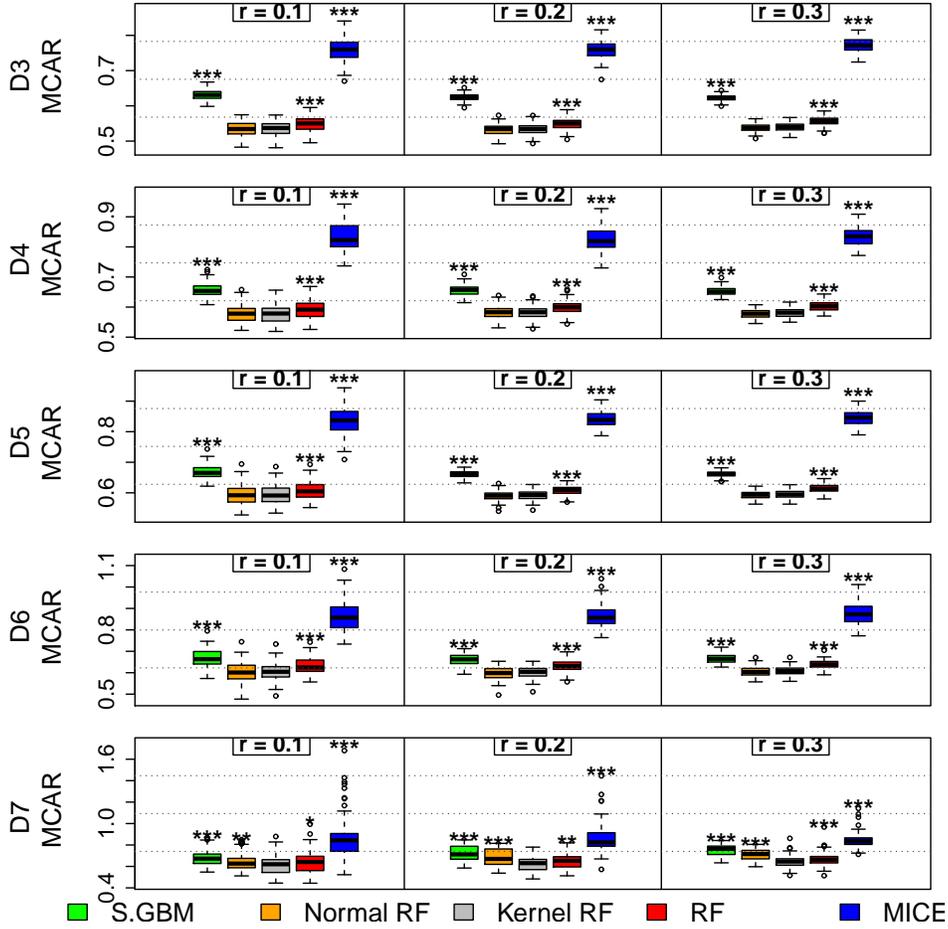}
		\caption{Imputation error measured by NRMSE under the \textbf{MCAR} scheme of the stochastic gradient tree boosting (S.GBM), the random forest with parametric boostrapping (Norm RF), the random forest with kernel sampling (Kernel RF), the (classical) random forest (RF) and the \texttt{mice} method (MICE) 	
			on data sets \textbf{(D3)} --  \textbf{(D7)}  with various missing rates ($r \in \{0.1, 0.2, 0.3\}$) . }
		\label{fig2}
	\end{figure}
	
	\begin{figure}[h]
		\centering
		\includegraphics[scale=0.7]{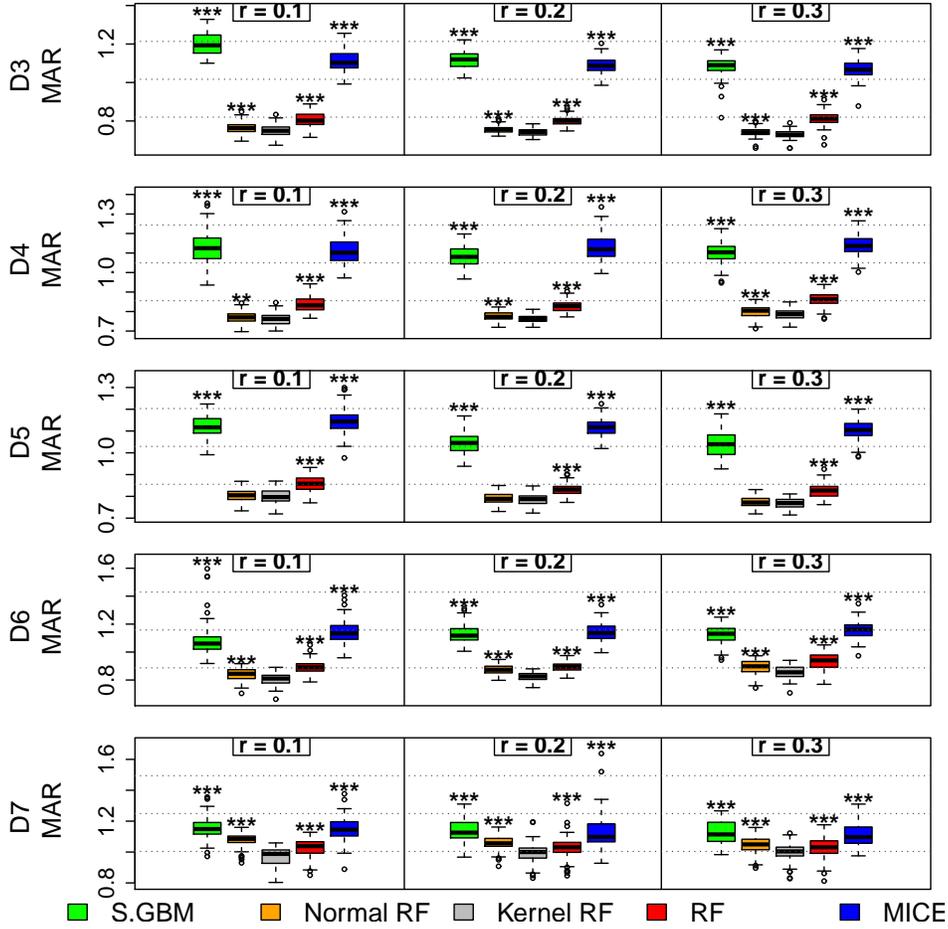}
		\caption{Imputation error measured by NRMSE under the \textbf{MAR} scheme of the stochastic gradient tree boosting (S.GBM), the random forest with parametric boostrapping (Norm RF), the random forest with kernel sampling (Kernel RF), the (classical) random forest (RF) and the \texttt{mice} method (MICE) 
			on data sets \textbf{(D3)} --  \textbf{(D7)} with various missing rates ($r \in \{0.1, 0.2, 0.3\}$). }
		\label{fig3}
	\end{figure}
	
	The results are displayed in Figure \ref{fig2} and  Figure \ref{fig3} for the MCAR and MAR mechanism, respectively. Imputing missing values in continuous outcomes using the S.GBM resulted in less accurate imputation results compared to all random forest methods. These findings were consistent through all data sets with continuous outcomes and under all $30$ simulation settings. From the RF methods, a random forest with resampling from a multivariate Gaussian kernel (Kernel RF), showed the 
	best overall results among all five methods compared in this paper. In particular, even when the assumption of multivariate normality was not met and substituted by heavy-tailed or non-symmetric distributions, the Kernel RF yielded favorable imputation accuracy. Only in case of MCAR and data sets \textbf{(D3)} --  \textbf{(D6)} the RF method based on parametric bootstrapping from an estimated multivariate normal distribution (Normal RF) showed comparable results. In all other settings, the Kernel RF resulted in the lowest imputation error. These findings were also significant at the $1\%$ level using the one-sided Brunner-Munzel test.  For example, mean NRMSE could be reduced by $3\%$ in data set \textbf{(D7)} under the MCAR mechanism with $30\%$ missing rate when using Kernel RF. In case of a heavy-tailed distribution like in data set \textbf{(D5)}, mean NRMSE could be reduced by up to $7 \%$ under the MAR mechanism with $30\%$ missing rate. Within each missing mechanism and simulation set-up, imputation accuracy decreased slightly with increasing missing rates $r \in \{0.1, 0.2, 0.3\}$. However, it seems that the novel RF methods (Kernel and Normal RF) are slightly less affected by the missing rate than simple resampling within the random forest protocol. Comparable to the categorical cases, MICE yielded high imputation errors under both missing mechanisms. 
	
	For mixed-type data, we finally combine the resulting best procedures from the two previous sections, i.e. we apply the S.GBM and the Kernel RF method in cases where classification resp. regression tasks are required to be solved. The resulting imputation algorithm (\texttt{Miss BooPF}) is analyzed in detail in the sequel.
	
	\subsection{Mixed-type data}
	
	In this section, we consider data matrices $\mathbf{Y}$ consisting of both, continuous and categorical outcomes. In addition to the generated synthetic data sets, we were interested in the accuracy of imputation techniques on real-life examples. Therefore, two data sets have been considered, which we summarize in the following: 
	
	\vspace{0.5cm}
	
	\textit{German Credit Data} was acquired from the UCI Machine Learning Repository. The primary goal of the data set was to classify credit applicants as credit-worthy or not by considering a set of attributes. The data set consists of $p = 20$ variables from $n = 1,000$ customers applying for credit at a German credit granting institution. Three variables have been identified as continuous outcomes, whereas $17$ are of categorical nature (ten ordinal and seven nominal variables are present). Although missing values are not present, the level outcome of 
	attributes $\# 12$ (property) and $\# 19$ (telephone) might be misleading. We interpret the feature characteristic 'unknown/no property' and 'none' as potential feature levels.    
	
		\vspace{0.5cm}
	
	\textit{myPersonality.com} was a Facebook application developed by David Stillwell from the University of Cambridge. In collaboration with more than 200 researchers worldwide, psycho-demographic profiles of more than $8$ million Facebook users were analyzed.  The application provides psychological questionnaires  and real psychometric tests while recording Facebook profiles from users across the world. For our simulation study, $n = 463$ Facebook profiles of German and British users were analyzed. We extracted $p = 13$ variables such as  \textit{gender, age, relationship status, interests, locality, number of friends, timezone, political attitude, religion, IQ score, life satisfaction, employment status} and \textit{account duration}. Some features such as \textit{political attitude} and \textit{religion} had diverse characteristics with a same or similar meaning. Therefore, we grouped \textit{political attitude} into levels \textit{apathetic, conservative, democratic, liberal }
	and \textit{others}. The variable \textit{religion} was grouped into \textit{Christian, Muslim} and \textit{Others}.
	
	\begin{figure}[h]
		\centering
		\includegraphics[scale=0.7]{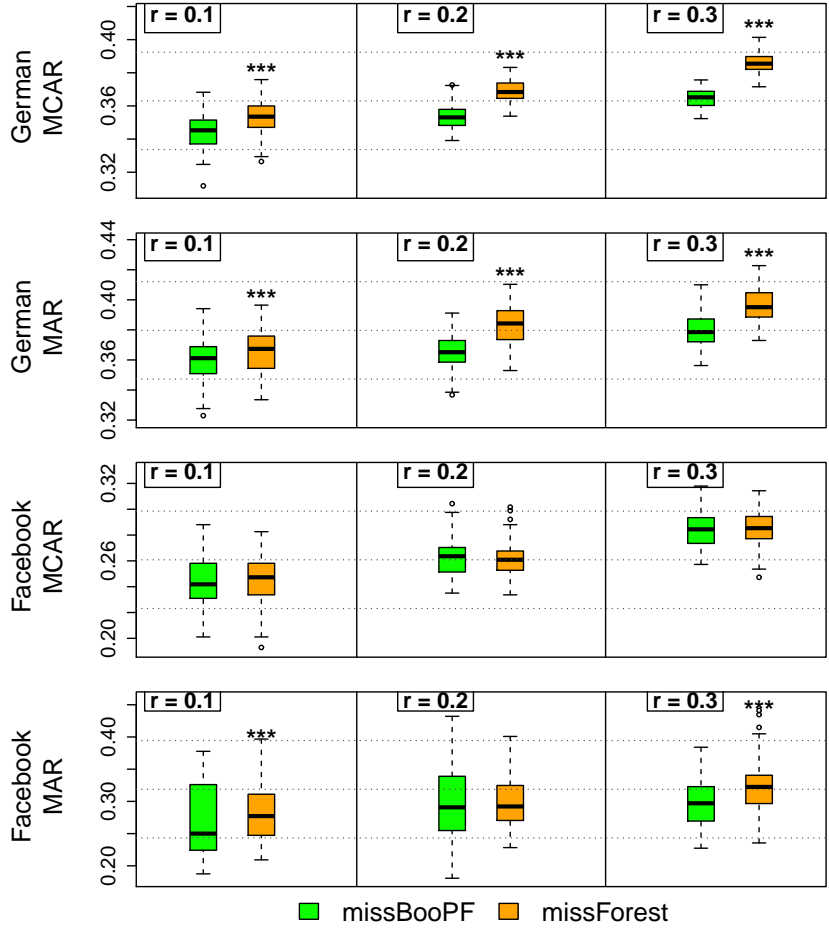}
		\caption{Imputation error measured by \textbf{PFC} of the combined imputation procedure using the gradient tree boosting and the parametrically bootstrapped random forest with kernel re-sampling (missBooPF) as well as the usual random forest method (missForest) on the \textbf{German Credit Data} and \textbf{Facebook Data} under various missing rates ($r \in \{0.1, 0.2, 0.3\}$) and mechanisms. }
		\label{fig4}
	\end{figure}
	
	\begin{figure}[h]
		\centering
		\includegraphics[scale=0.7]{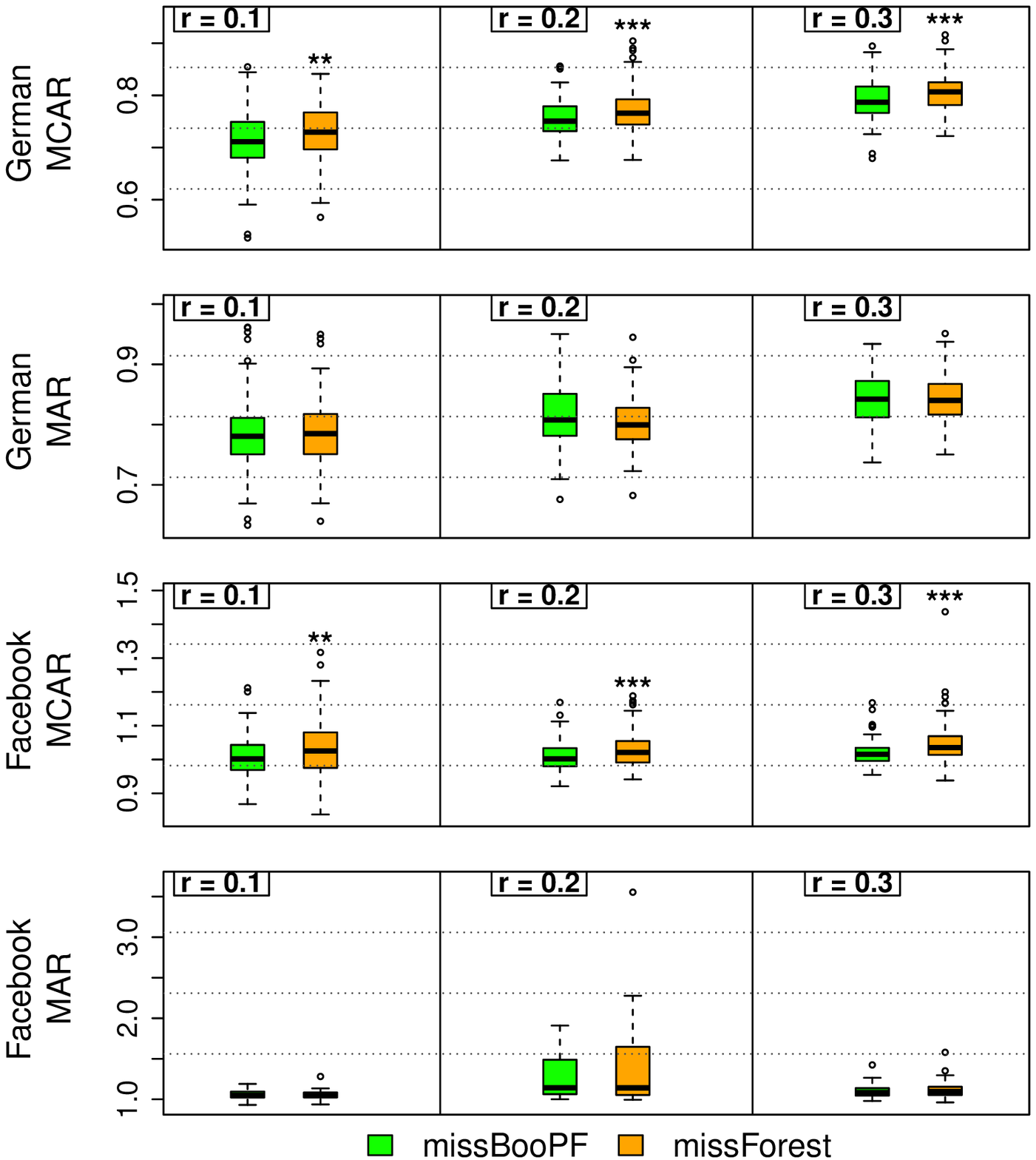}
		\caption{Imputation error measured by \textbf{NRMSE} of the combined imputation procedure using the gradient tree boosting and the parametrically bootstrapped random forest with kernel re-sampling (missBooPF) as well as the usualy random forest method (missForest) on the \textbf{German Credit Data} and \textbf{Facebook Data} under various missing rates ($r \in \{0.1, 0.2, 0.3\}$) and missing mechanisms. }
		\label{fig5}
	\end{figure}

	\vspace{0.5cm} 
	The simulation design for mixed-type data remains the same as before. Under the two missing mechanisms MCAR and MAR with various missing rates the best procedures from Section 5.1 and 5.2 are chosen and applied on the German Credit Data as well as on the \textit{myPersonality.com} data set. In particular, in case of classification, S.GBM is chosen and for regression, we took advantage of the parametric version of the random forest using the kernel method with the normal scale rule (Kernel RF). We denote the combination as \texttt{Miss BooPF} and compare its imputation accuracy with that of the missForest method. 
		\vspace{0.5cm}
		
	The results are displayed in Figure \ref{fig4} (categorical outcomes) and Figure \ref{fig5} (continuous outcomes) and are similar to the one obtained from synthetic data generation. In particular, the \texttt{Miss BooPF} increases the imputation accuracy considerably in most scenarios, where its advantage is slightly more pronounced in the MCAR cases. Considering the Facebook data set, mean NRMSE could be reduced by around $2 \%$ under the MCAR mechanism with $20\%$ missing rate. For the German Credit data set, mean PFC could be reduced by up to $5 \%$ under the MAR mechanism with $20\%$ missing rate.
		\vspace{0.5cm}
		
	Finally, note that the effect of the Kernel RF procedure is less distinct due to the relatively low number of continuous outcomes in both data sets.
	
	\newpage
	
	\section{Conclusion}
	We proposed several modifications of the random forest method and investigated its performance for imputing missing values following the proposal of \cite{stekhoven2011missforest,stekhoven2011using}. The key idea was to change the bootstrap step within the random forest by means of other resampling techniques. Imputation accuracy was measured for MCAR and MAR under various settings. In particular, continuous, categorical as well as mixed-type data were analyzed. 
	
	In case of only continuous variables, the proposed parametric random forest method based on kernel sampling (Kernel RF) yielded favorable results in terms of higher imputation accuracy. This result could be observed under the MCAR and MAR conditions across the various missing rates of $0.1$, $0.2$ and $0.3$. Even under the normal scale rule for choosing an optimal bandwidth matrix, violations towards the assumptions of normality did not affect the imputation accuracy dramatically. Instead, it improved existing imputation procedures such as \texttt{mice} and \texttt{missForest} in the synthetic data set-up as well as in the empirical analysis for mixed-type data. 
	For categorical variables, the stochastic gradient tree boosting (S.GBM) performed comparably well and yielded better results than the \texttt{missForest} imputation method when level frequencies are not equally present. The same observation was met for other competitors such as mice or random forest procedures based on other resampling strategies such as stratified sampling. 
	Based on these findings we proposed to apply a mixture of the Kernel RF and the S.GBM depending on the regression resp. classification task. The resulting \texttt{Miss BooPF} imputation method is applicable to all kind of data. In particular, its applicability for mixed-type data was exemplified by additionally imputing data from empirical studies on credit information and Facebook data. Again \texttt{Miss BooPF} yielded improved results for both, continuous as well as categorical outcomes. 
	
	Due to the increased imputation accuracy of the Kernel RF method, we plan to study its practical performance in pure regression problems together with theoretical analyses (such as consistency in metric spaces).  
	
\section*{Acknowledgement}
We are thankful to David Stillwell from Cambridge University and Michal Kosinsky from the Stanford Graduate School for providing us the Facebook data. We appreciate the initial support of the Daimler AG represented by the research department 'Computer Vision and Camera Systems'
	
	
	\newpage
	
	\appendix
	\section*{Appendix A.}
	Missing values have been inserted artificially under consideration of the various missing mechanisms. For the MCAR, MAR and MNAR condition, a special kind of mechanism has been implemented, which will be described in the following:
	
	\begin{enumerate}
		\item \textbf{Missing Completely at Random.} We replace values randomly with missing values. For every variable $\mathbf{X}_{j}$, $j = 1,...,p$, we assumed that $R_{ij} \stackrel{iid}{\sim} Bernoulli(1-r)$, $i = 1,...,n$ where $r \in \{0.1, 0.2, 0.3\}$ is the overall missing rate based on $n \cdot p$. 
		\item \textbf{Missing at Random.} We implement this mechanism by building dependency structures across missing values of subsequent variables using the logistic regression. First, randomly select $j^{* }\in \{1,...,p\}$ as the initial index and assume that $R_{ij^{*}} \stackrel{iid}{\sim} Bernoulli(1 - r)$, where $r \in \{0.1, 0.2, $ $0.3\}$ is the overall missing rate. The missing values for the subsequent variable $ \mathbf{X}_{j_{s}^{*}}$ are inserted using the observed components of $\mathbf{X}_{j^{*}}$ as covariate values within a logistic regression model. The response variables are randomly generated in an upstream step to estimate model parameters. Therefore, let $\mathbf{X}_{j^{*}}^{obs}$ be the sub-vector of observed components of $\mathbf{X}_{j^{*}}$. We construct a training response by generating $\tilde{R}_{ij_{s}^{*}} \stackrel{iid}{\sim} Bernoulli(1 - r)$ for all $i \in \mathbf{i}_{j^{*}}^{obs}$ and set $\{ \mathbf{\tilde{R}}_{j_{s}^{*}}, \mathbf{X}_{j^{*}}^{obs} \}$ as the training sample on which the logistic regression will be conducted. If $\hat{p}_{i j_{s}^{*}} $ is the predicted probability of $(\tilde{R}_{ij_{s}^{*}} = 1 | X_{ij^{*}}),$ $i \in \mathbf{i}_{j^{*}}^{obs} $, then for the observations $k \in \mathbf{i}_{j^{*}}^{obs}$ with the $\lfloor r \cdot n \rfloor$ - smallest values of $\hat{p}_{i j_{s}^{*}}$, we set  $R_{kj_{s}^{*}} = 0$. The process is continued in a pairwise fashion until all variables have been treated. 
		
		\item \textbf{Missing Not at Random.} We implement a specific type of the MNAR mechanism by constructing censored data. For every column vector $\mathbf{X}_{j}$, $j \in \{1,...,p\}$, sort the observations in increasing order and randomly select an observational point $x_{(i)j}$, $i \in \{1,...,n\}$ of the ordered sequence $\mathbf{X}_{(j)} =(X_{(1)j},X_{(2)j},...,X_{(n)j}  )^{\top}$. We denote with $\pi_o$ the resulting order-permutation. Depending on whether $x_{(i)j}$ lies above or below its $r-th$ resp. $(1-r)-th$ quantile, missing values are inserted, i.e. $R_{kj} = 0$ for
		$$
		k \in \begin{cases}
		L_{j} & \text{if } x_{(i)j} < x_{(r)j}, \\
		U_{j} & \text{if } x_{(i)j} > x_{(1-r)j}, \\
		B_{j} &\text{else, } 
		\end{cases}
		$$
		
		with $ L_{j} = \{ \pi_o(i),\pi_o(i+1),\dots,\pi_o(i+\lfloor r\cdot n \rfloor -1) \}$, $ U_{j} = \{ \pi_o(i),\pi_o(i-1),\dots,\pi_o(i-\lfloor r \cdot n \rfloor +1) \}$ and $ B_{j} = \{ \pi_o(i), $ $\pi_o(i +\varphi(U)\cdot1),\dots,\pi_o(i+\varphi(U) \cdot(\lfloor r \cdot n \rfloor -1)) \}$ such that $\varphi(x) = \mathds{1}_{[0, 0.5)} - \mathds{1}_{[0.5, 1)}$, $U \sim Unif(0,1)$. 
	\end{enumerate}
	
	\newpage
	\bibliographystyle{ieeetr}

\end{document}